# AN IMAGE-BASED SENSOR SYSTEM FOR AUTONOMOUS RENDEZ-VOUS WITH UNCOOPERATIVE SATELLITES

C. Miravet[1], L. Pascual[1], E. Krouch[1], J. M. del Cura[1]

[1]SENER, Ingeniería y Sistemas, c\ Severo Ochoa 4, PTM, Tres Cantos, Madrid, SPAIN
carlos.miravet@sener.es

## INTRODUCTION

In this paper are described the image processing algorithms developed by SENER, Ingeniería y Sistemas to cope with the problem of image-based, autonomous rendez-vous (RV) with an orbiting satellite. The methods developed have a direct application in the OLEV (Orbital Life Extension Extension Vehicle) mission. OLEV is a commercial mission under development by a consortium formed by Swedish Space Corporation, Kayser-Threde and SENER, aimed to extend the operational life of geostationary telecommunication satellites by supplying them control, navigation and guidance services. OLEV is planned to use a set of cameras to determine the angular position and distance to the client satellite during the complete phases of rendez-vous and docking, thus enabling the operation with satellites not equipped with any specific navigational aid to provide support during the approach.

The ability to operate with un-equipped client satellites significantly expands the range of applicability of the system under development, compared to other competing video technologies already tested in previous spatial missions, such as the ones described here below.

**Videometer**

Used by the Jules Verne ATV (Automated Transfer Vehicle). The ATV vehicle was conceived as a resupply and reboost vehicle capable of performing a fully automated docking with the International Space Station. The ATV, which has recently performed its first mission with remarkable success, uses as primary sensor at close-ranges videometers to determine its distance and orientation relative to its target. Based on the design of a star tracker, the videometer is the first automatic optical operational system ever used for spacecraft navigation. It operates by emitting a pulsed laser beam, which is reflected by an arrangement of passive retro-reflectors (26, in the ISS) installed in the client satellite. Analysis of the time of flight and the pattern of reflected light enables to determine the distance and direction to the docking port. To add redundancy and a safety margin to the critical rendez-vous operations, the relative distance and orientation to the client satellite is also computed by a secondary independent sensor, a telegoniometer.

**AVGS (Advanced Video Guidance Sensor)**:

Another video sensor for rendez-vous and docking operations is the AVGS (Advanced Video Guidance Sensor), used in the Dart and Orbital Express demonstration missions. Dart was developed as part of NASA's Space Exploration Initiative, and was launched on April 15, 2005, aboard an Orbital Sciences Pegasus Launch Vehicle. The Dart mission included performing autonomous proximity operations on a target satellite using as primary sensor the AVGS. DART performed as planned during the first eight hours through the launch, early orbit, and rendez-vous phases of the mission, accomplishing all objectives up to that time. However, during proximity operations, the spacecraft began using much more propellant than expected. As DART detected its propellant source was approaching exhaustion, it began a series of manoeuvres for departure and retirement. DART finally collided with its client satellite 3 minutes and 49 seconds before initiating retirement. The Orbital Express mission was funded by DARPA and NASA with the goal of validating the technical feasibility of robotic, autonomous on-orbit refuelling and reconfiguration of satellites. Orbital Express consists of a robotic servicing spacecraft (ASTRO) and NextSat, a prototypical modular next-generation serviceable client spacecraft. The tests performed in orbit included connection through the robotic arm, fuel transference, formation flying and autonomous rendez-vous, and were completed successfully.

The AVGS system combines an imaging sensor, integrated laser sources, and narrow-band-filtered retro-reflective targets with sophisticated signal processing and optical correlation to develop six-degree-of-freedom estimates of the relative state between two spacecraft engaged in proximity operations. The AVGS sensor is equipped with four multiplexed 1 watt diode lasers illuminating a 24º aperture in front of the target vehicle. Laser light is reflected by reflective corner cubes mounted on the target satellite, and arranged in a triangle pattern with the optical axis of all three corner cubes pointing in the same direction. The reflected light is collected by an imaging lens and projected onto a 1024 x 1024 CMOS detector. Pattern recognition algorithms identify the spot centroids of each corner cube to determine range, bearing, and attitude of the chaser vehicle with respect to the target satellite. The use of sensors such as the videometer or the AVGS obviously requires the previous installation of the set of retro-reflectors on the client satellite. This precludes their use in a mission such as OLEV, conceived to operate with satellites already in orbit, and generally un-equipped with any specific means to ease the rendez-vous and docking operation.

**Radiofrequency**

Radiofrequency (RF) has also been used in space missions to sense distance and direction to a given target. An especially relevant example based in RF sensors is the Russian *Kurs* system, which has been used for rendez-vous navigation for a long time. This system was designed to provide all required navigation measurements during the entire approach from a few hundreds of kilometres down to contact, and operates in the S-band, using wavelengths of the order of 10 cm in continuous and modulated signals. In the RF approach, a modulated RF signal is generated by a transmitter and is directed by an antenna towards the target. Part of the signal's power is reflected by the target back in the direction of the transmitter, or the signal is re-transmitted by a transponder at the target, and is received by the antenna at the transmitter's location. The typical maximum range of RF docking sensors is on the order of 100 Km, fitting well with the long and medium range of the approach, whereas the very short range would require a substantial effort to mitigate disturbances and provide the required performance. Other important drawbacks of the RF technology are the substantial power consumption and mass of the antennas required to operate over long distances to the target, making this choice of sensor less used in modern systems.

**The LAMP (LAser MaPper)**:

Used in the ST6 NASA mission launched in early 2004, this sensor is a laser radar. It operates by emitting short high power laser pulses, which bounce off an internal gimballed mirror that determines the azimuth and elevation of the outgoing beam. When the laser pulse hits a target, a small amount of the light is reflected back to the instrument. The returned laser pulse bounces off the internal mirror and is collected by a telescope. On the way out, a laser actuated trigger starts a counter that is stopped by the return pulse. By sweeping the internal gimballed mirror though a number of angles, it is possible to form a 3-dimensional image of the space in front of LAMP. The laser beam has been shaped to have a 0,02° (0.35 mrad) divergence. For a target surface with a 0.1m area, the detection range is 2.5 km. The LAMP telescope itself has a 5 cm aperture and is a classical Cassegrain type.

**Light Detection and Ranging (Lidar)**

This is an optical remote sensing technology that measures properties of scattered light to find range and/or other information of a distant target. The prevalent method to determine distance to an object or surface is to use laser pulses in a similar way to radar technology, which uses radio waves instead of light. The range to an object is determined by measuring the time delay between transmission of a pulse and detection of the reflected signal. LIDAR enables technology for Lunar Science, Exploration and Resource Prospecting. For in space and on orbit rendez-vous, the powerful flash LIDAR LASER pulse can be used to acquire and range targets from a distance of up to 10-20 km. For docking applications, flash LIDAR provides real-time three dimensional video of the target spacecraft under any lighting conditions. This provides six degree of freedom pose as well as velocity and spin rate data. Inclusion of a flash LIDAR system also allows for redundant video guidance capabilities.

There have been other missions (like Rosseta ESA's mission, Gemini or NEAR NASA's mission or Japan's Hayabusa mission) trying to reach passive (non cooperative) targets in space such as comets, planets, etc. Most of them used cameras as one of the main optical instruments to successfully perform the rendez-vous, but supporting it with secondary optical instruments like a laser range finder to get distance measurements. In Hayabusa mission, the navigation and guidance method was based on a camera, extracting images and processing them by criteria based on their spatial wavelength and variance of their brightness. Based on these criteria, the autonomous system was able to extract three fixation points with enough contrast to be clearly identified and tracked, allowing the chaser vehicle to widen the applicable range of the camera even when the target goes out of the field of view. The fixation points were extracted processing an image applying least-square block matching to a window of variable size, using band pass filters and variances in this window. The high local variances were used as templates to do the tracking. These templates could be selected in terms of a precision or a roughness criterion dependent on the window size.

In summary, use of passive video cameras to sense direction and distance results in distinct advantages in a mission such as OLEV, which has to interact with satellites already in orbit and not equipped with navigational aids to ease docking, and where low mass and power consumption is a primary requirement. In the following sections are briefly presented the main parameters of the RV phase of the OLEV mission, and the characteristics of the image processing module to accomplish the rendez-vous operation.

**OLEV RENDEZ-VOUS STRATEGY**

The OLEV satellite will firstly be left on a geostationary transfer orbit by a proper launcher (Ariane 5, PSLV-XL or LM-3). Then, under nominal operational conditions, it will begin the transfer to a geostationary orbit, using a low thrust electric propulsion system. The rendez-vous phase (see fig. 1) starts at point S1, 35 Km in front of the client. A radial impulse manoeuvre leads OLEV to point S2, 2 Km behind the client. At this point begins the image-based relative navigation, which provides sufficient precision to accomplish a safe approach to the client. The distance between both spacecrafts is first reduced to 100 m by a forced translation, propelled by its cold-gas thrusters and actively controlled. The OLEV reaches the next hold point at 50 m from the client with a fly-around manoeuvre and rendez-vous ends with a last forced translation until docking.

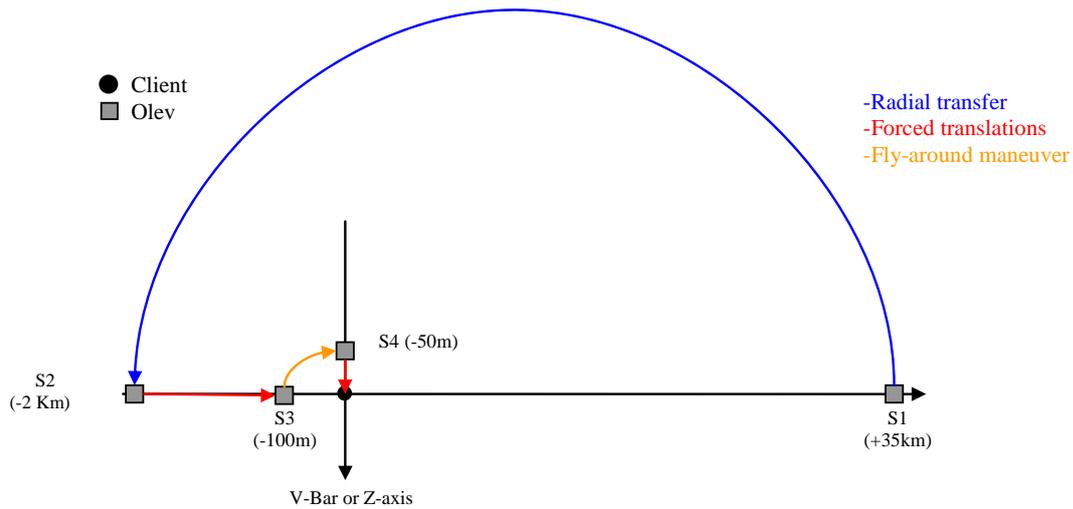

Fig 1. Phases of the rendez-vous operation, covering the approach to the target satellite from 35 Km (S1) to 5 m, where control is handled to the docking system. Image-based relative navigation starts at S2.

Special care has been taken in planning an approach procedure that ensures a correct illumination of the client satellite by the Sun during the complete RV phase. In fig. 2 are presented the solar light incidence angles during the complete RV manoeuvre, ranging from 9º at the start of the RV to 65º when control is handled to the docking system. In particular, this trajectory precludes the possibility of a direct viewing of the Sun during the RV manoeuvre.

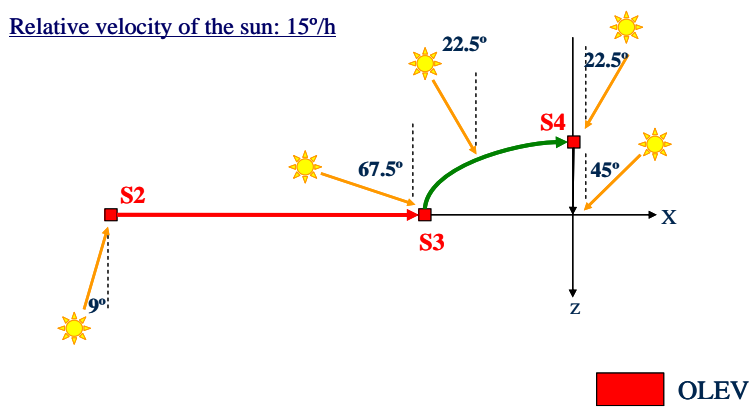

Fig 2. Sun illumination angles during the OLEV's RV manoeuvre.

The RV imaging system is composed of two sets of cameras, each set consisting of identical nominal and redundant units, to assist respectively on far (2 Km – 100 m) and mid (100 m – 5 m) range rendez-vous operations. Both the far and mid range cameras are based on the STAR 250 radiant tolerant APS, with an image size of 512x512 pixels. For each range, the camera focal length is selected to fulfil the following conditions:

- the client satellite image (or the part of it selected for autonomous detection/tracking) at the farthest limit of the range is large enough not to be confused with point-like objects of the stellar background.
- the client satellite image (or the part of it selected for autonomous detection/tracking) at the closest limit of the range is small enough to allow complete viewing in the camera's field of view (FoV) with a safety margin of around half of the image.

In fig.3 are presented simulated views of a client satellite, as seen by the far and mid range cameras at different steps of the RV manoeuvre.

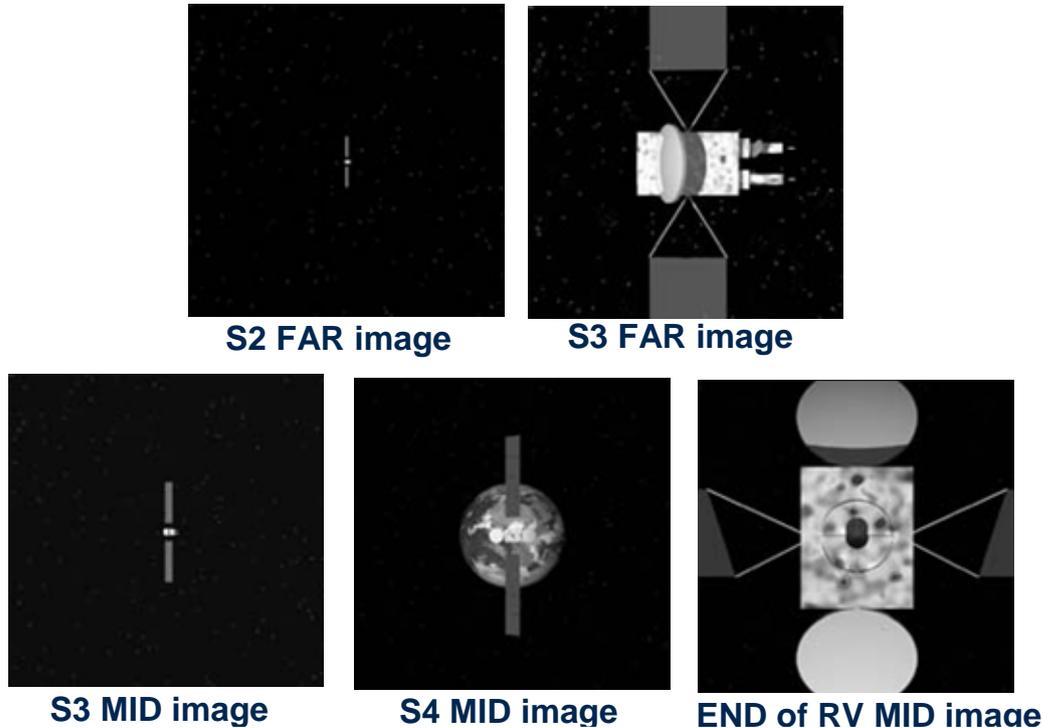

Fig 3. Simulated images of the client satellite during the RV manoeuvre.

The images captured by the active RV camera (far/mid range) are downloaded to ground and automatically processed to obtain a quasi-real time estimation of the angular position and distance to the client satellite during the RV operation. The complete cycle of image downloading, processing and transmission of the determined angular position and distance to the OLEV control system is scheduled to last a maximum of 1 second, which sets an important limitation to the type of image processing algorithms that could be used, even if the processing is performed with on-ground resources. In addition, the image processing algorithms have to deal robustly with factors inherent to the operational scenario, such as noise, presence of a stellar background, variations in illumination and sometimes a considerably cluttered background, caused by the appearance of bright objects (Earth, Moon) in the camera's FoV. Taken as a whole, the automated processing of these images is a challenging task which is critical in terms of ultimate mission success. To cope with these problems, SENER has designed an image processing chain based on the use of morphological gray-filters by reconstruction [1-4], which have proven an excellent reliability and performance in environments as demanding as that of automatic mine detection [5], and had been successfully used by SENER on automatic airborne inspection of electrical power lines. Tests performed on actual imagery of rendez-vous operations (Jules Verne ATV, Orbital Express) have shown the excellent robustness and accuracy of the developed method, which is briefly outlined in the following section.

**OLEV IMAGE PROCESSING MODULE**

The image processing module of the OLEV operates processing on-ground the images downloaded from the satellite at 1 s intervals. The main functionalities of the image processing module are:

- **Automous Satellite detection**
  This function is used at the beginning of the RV manoeuvre, to detect and extract the shape of the client satellite in an image captured by the far RV camera. The extracted shape location, size and attitude are used to initialize the tracking procedure, which follows the target with sub-pixel precision during the approaching manoeuvre. The detection procedure could also be applied periodically during tracking to obtain an independent estimation of the satellite location parameters, for validation purposes.

- **Model-based satellite image tracking**
  Once the location of the client satellite is determined by the detection function, control is transferred to tracking, which uses a wireframe model of the satellite to determine its location in the image with sub-pixel precision. The model is translated, rotated and scaled in the framework of an optimization procedure, to obtain the best possible matching with the perceived contours of the satellite in the image.

- **Sub-pixel determination of satellite location parameters**
  From the parameters (translation, rotation, scaling) of the best fitting model, the angular position, range distance and relative attitude to the client satellite are determined. The determination of the best-fitting model transformation parameters with sub-pixel precision is important to ensure an adequate accuracy in the derived parameters. Particularly, this is the case of range determination from image scale when observing from the distant limit of the operational range of a RV camera. At this distance, the client satellite image spans a few pixels, with a large associated quantization error if image scale is determined with accuracy at just the pixel level.

The procedures used to perform these functions, with examples of operation of real RV imagery, are presented in the following sections.

**Autonomous Satellite Detection**

The basic aim of the detection module is to recognize the presence of the client satellite in the incoming image and the extraction of its shape with sufficient accuracy to enable the initialization of the tracking mode. A basic desirable characteristic of the detection module is robustness against factors such as noise and background clutter, as initial correct localization of the satellite shape on the image is vital for ultimate mission success. In fig. 4 is presented an image obtained during the Orbital Express demonstration mission, where the client satellite is observed at mid-distance surrounded by a considerable amount of background clutter mainly caused by the texture of the cloud pattern. In the figure is presented the image histogram, with two clear lobes, corresponding to 'cloud' pixels (bright lobe) and 'sea' pixels in the upper left side of the picture (dark lobe). The distribution of the satellite pixel gray levels is placed somewhere in between these two main lobes, as can be observed in the satellite pixel histogram included in the figure. As the results of any automatic binarization procedure will be driven by the position of these two main lobes, the satellite pixels could be alternatively assigned to the target or the background, resulting in a poor or at least unreliable segmentation of the image for our purposes. To be able to process robustly images of this complexity, SENER has adopted an alternative approach, based on the use of morphological filters [1, 2]. These filters enable the robust extraction of shapes that present a contrast with the local background (either positive or negative), and are totally enclosed by a window of given size. In fig. 4 are presented the result of filtering the input image with a dilation filter of size 17 pixels. This filter substitutes each pixel in the image by the maximum value in a local neighbourhood defined by the filter window size, completely removing dark objects smaller than this size. Unfortunately, this operation introduces a significant distortion in the rest of the image structure, hindering the detection of the completely removed regions. As could be seen, the situation is only partially alleviated by applying an erosion, the inverse operator to dilation, to obtain what is known as a morphological closing. On the contrary, results are highly improved by applying after dilation an erosion by reconstruction filter instead of basic erosion, to obtain a closing by reconstruction [3, 4]. As could be observed in the figure, this operation performs a pretty fine job in removing dark objects below the specified size without severely affecting bright objects, or dark objects above that size.

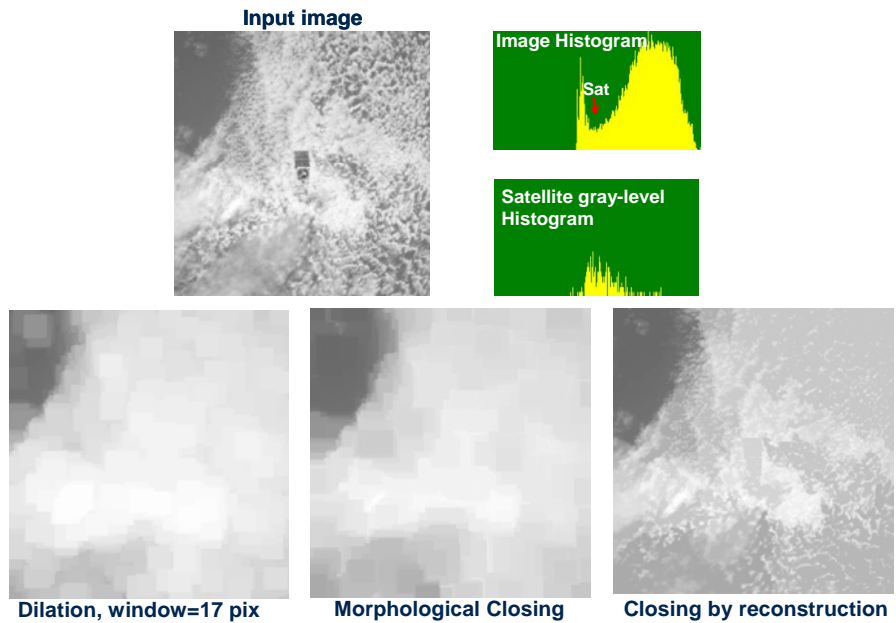

Fig. 4. Top: input image to the detection module (Orbital Express mission); image histogram and histogram of satellite pixels gray level. Bottom: effect of several morphological filters on the input image (see text).

The image obtained by the closing by reconstruction filter could be taken as a background image, where all potential objects of interest have been removed. Subtracting from this image the input data, we obtain the results of an operation known as top-hat closing filtering by reconstruction, which here highlights satellite pixels together with those pixels in the background fulfilling the same constraints in local contrast sign and shape size. The results of the top-hat filtering are shown in the central panel of fig. 5.

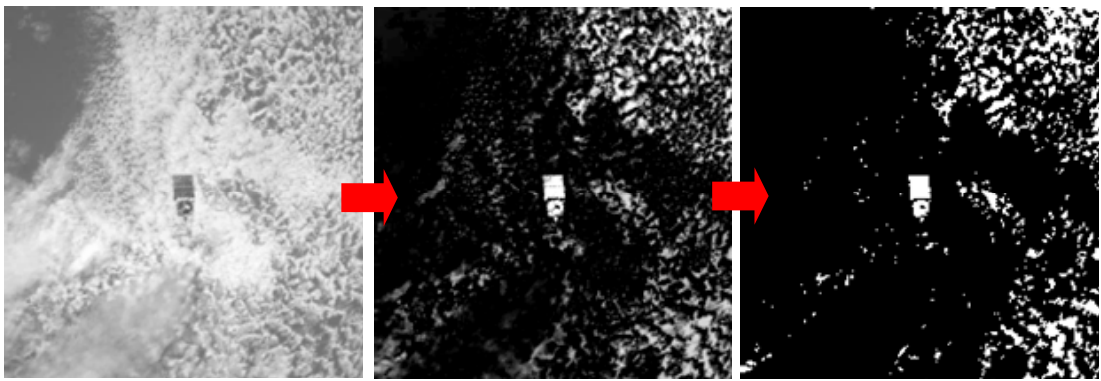

Fig. 5. Input image (left); results after top-hat closing filtering by reconstruction (center); results after image binarization (right).

The result of the morphological filtering is still a gray-level image, which is assumed to be constituted by a mixture of pixels belonging to two classes:

- *Target class*: formed (in this example) by pixels belonging to objects in the scene that present a negative contrast with respect to the background and are smaller in size than the applied filter window. Ideally, this class will comprise pixels contained in the satellite shape, together with pixels of other objects in the background fulfilling the same criteria.

- *Background class*: rest of image pixels, belonging to objects in the background not fulfilling the extraction criteria.

Pixels in the target class are enhanced by the morphological filtering operation and, hence, will appear in principle in the upper part of the gray-level histogram, whereas the background pixels will form in the histogram a large lobe close to the origin. In fig. 6 (left panel) is presented the histogram of the top-hat filtered image, where this hypothesis is fulfilled. Assuming image pixels are constituted by two classes with separate pixel distributions, both classes could in principle be discriminated using a binarization procedure. In this work, the performance of several automatic thresholding methods based on different principles [6] have been assessed on a limited database of images obtained from actual space RV missions (Orbital Express, ATV). Preliminary results point out the superiority for this application of thresholding algorithms based on maximization of entropy–related criteria.

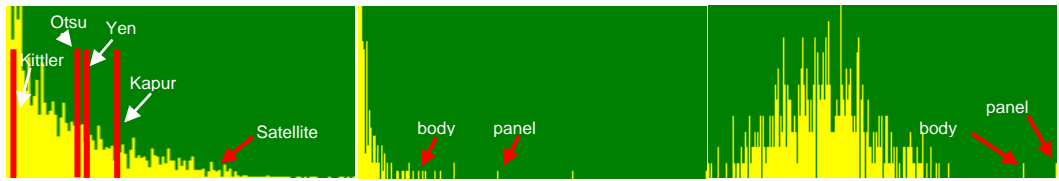

Fig. 6. Histogram of the top-hat filtered image, with thresholds selected by different binarization algorithms (left). Histograms of region attributes after binarization: area (center); local contrast (right).

After binarization, regions formed by connected groups of 'target' pixels are extracted by application of a classical labelling procedure [7]. These regions are then filtered using several criteria to separate regions corresponding to the spacecraft from background clutter. In this work, filters based in area, local contrast and shape have been considered. In fig. 6 are shown the histogram of region's area (central panel) and local contrast (right panel) with the position occupied by the regions corresponding to the body and solar panel of the client satellite. As can be observed, the area of the extracted regions is highly concentrated around the origin, with few regions with areas close to those expected for both the satellite body and solar panel. In the right panel of fig. 6 is presented the histogram of region's local contrast, defined here as the absolute difference between the average gray level of a region and that of its surrounding neighbourhood, measured on the filtered image. The histogram shows in the left a bell-shaped distribution corresponding to regions in the background clutter. The satellite body and panel regions appear distinctly on the right of the histogram, due to the higher local contrast. These results highlight the possibility of filtering out the spacecraft components using a combination of just size and local contrast criteria. However, in order to further increase the robustness of the filtering step, the approach was taken of applying these filters with very conservative ranges of both size and local contrast, and supplementing the filtering with a last stage that examines the shape of the selected regions. In the central image of fig. 7 are shown the regions selected after applying area and local contrast filters with fairly loose parameter ranges. In the right image, are shown the final results after filtering in terms of the shape of pre-selected regions.

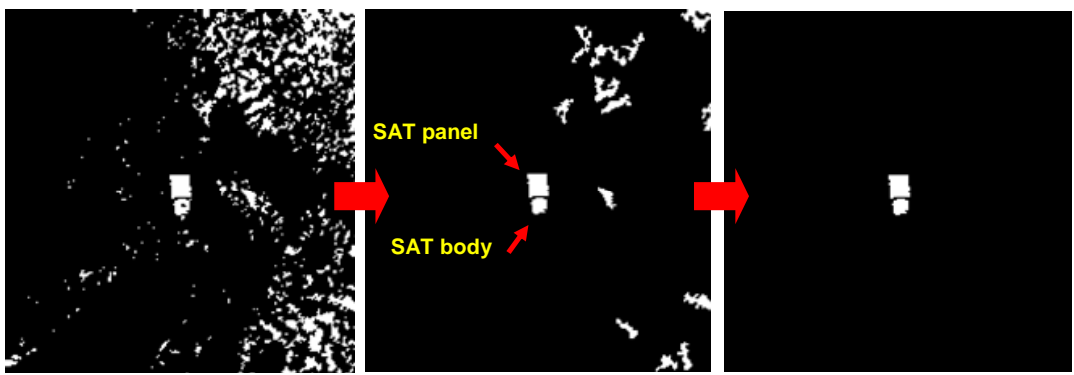

Fig. 7 Binarization results (left); regions pre-selected after filtering using a combined area and local contrast criterion (center); final regions selected after shape analysis (right).

In this case, the main components of the target spacecraft are known to present a circular (satellite body) and a rectangular shape (solar panel) as seen from the approaching trajectory. Hence, pre-selected regions are evaluated using a measure of circularity, such as compactness, and a measure of rectangularity, such as the ratio of the region's area to that of the minimum bounding rectangle. In fig. 8 are presented the values of these attributes for the spacecraft components and for several regions of the background. The significant difference in feature values for both classes confirms the possibility of performing a final reliable filtering stage based on this criterion.

| | | | | | |
|---|---|---|---|---|---|
| **Compactness (circularity)** | 0.96 | 0.80 | 0.31 | 0.32 | 0.69 |
| **Area/MBR_area (Rectangularity)** | 0.78 | 0.85 | 0.33 | 0.43 | 0.40 |

Fig. 8. Values of attributes measuring circularity (compactness) and rectangularity (ratio of area to that of the minimum bounding rectangle) for the spacecraft components and several background regions.

Finally, in fig. 9 are presented the results of applying the described detection algorithm to imagery captured during the course of ESA's ATV [8] rendez-vous and docking manoeuvre. In the first column are presented several input images. In the second column are shown the results of morphological processing. Finally, the third column shows the results after region filtering using a combined criterion of just area and local contrast. As can be observed, the detection results are excellent even in cases, such as that of the image presented in the second row, where the target (the ATV, in this case) is barely visible by visual standards. The images to be processed during the OLEV's rendez-vous will resemble most of the time the situation presented in the image in the first row, with the target object isolated in a stellar background.

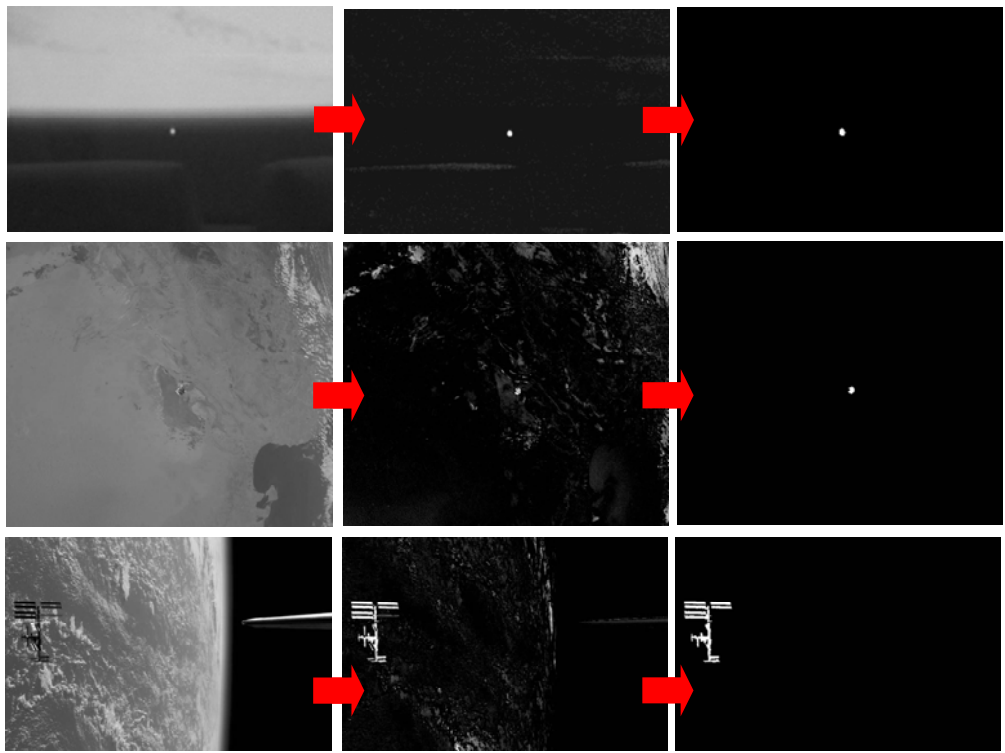

Fig. 9. Automatic detection of the target satellite on imagery captured during the ESA's ATV rendez-vous manoeuvre with the International Space Station. First column: input images; second column: results after morphological processing; third column: results after region filtering based on a combined area and contrast criterion.

**Model-based image tracking and parameter determination**

Once the image of the client satellite has been detected, the control is handled to the tracking module, which uses a simplified model of the object to follow its evolution during the video sequence of the approach. In fig. 10 is presented an outline of the procedure. First, a region of interest around the location of the target in the previous frame is processed using the morphological filter. The contours of the filtered image are then extracted using an edge detector [9]. At his point, a simplified wireframe model of the client satellite in projected onto the image, with parameters given by the results of the tracking procedure in the previous sequence image. In case of the 2D model used in this paper, four parameters are required to project the model onto a given sequence frame:

- Position ($x$, $y$) of a reference point in the model
- Model scale
- Orientation angle

A figure of merit of the alignment between model and image is computed in terms of the degree of matching between projected model lines and image contours. A numerical optimization process using the simplex downhill algorithm is carried out in the parameter space to bring this alignment measure to a local maximum. The optimal projection parameters (position, scale, angle) provide the necessary information to compute the angular position of the target and its distance and orientation relative to the chaser vehicle.

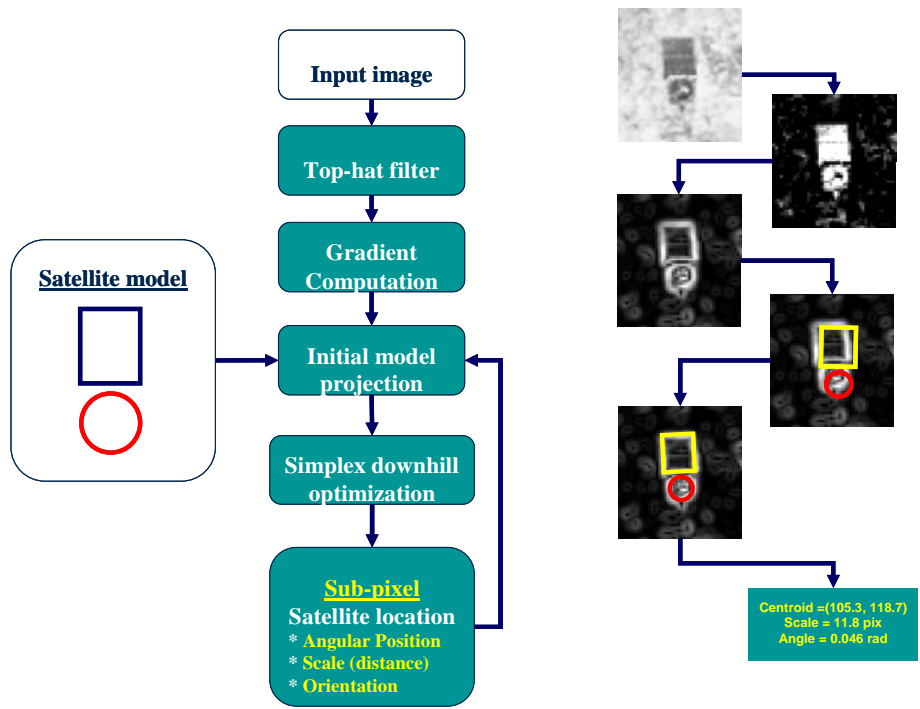

Fig. 10. Outline of the model-based image tracking and satellite parameter determination procedure.

In fig. 11 are presented some frames showing the results of tracking on a sequence obtained from the Orbital Express mission. In the upper-left side of each frame is presented a frame of the original sequence. Below it, the results of morphological processing and, at the bottom left, a graph showing relative distance to the target as computed from the perceived target scale. At the right side of each frame is presented an enlarged view of the target zone in each sequence with the wireframe model superimposed at the best fitting position. Tracking results were found to be robust and visually accurate despite the inaccuracies of the used satellite model and the vastly changing background, including frames in which the target show a very low contrast with the background (see, for instance, the central frame in fig. 11).

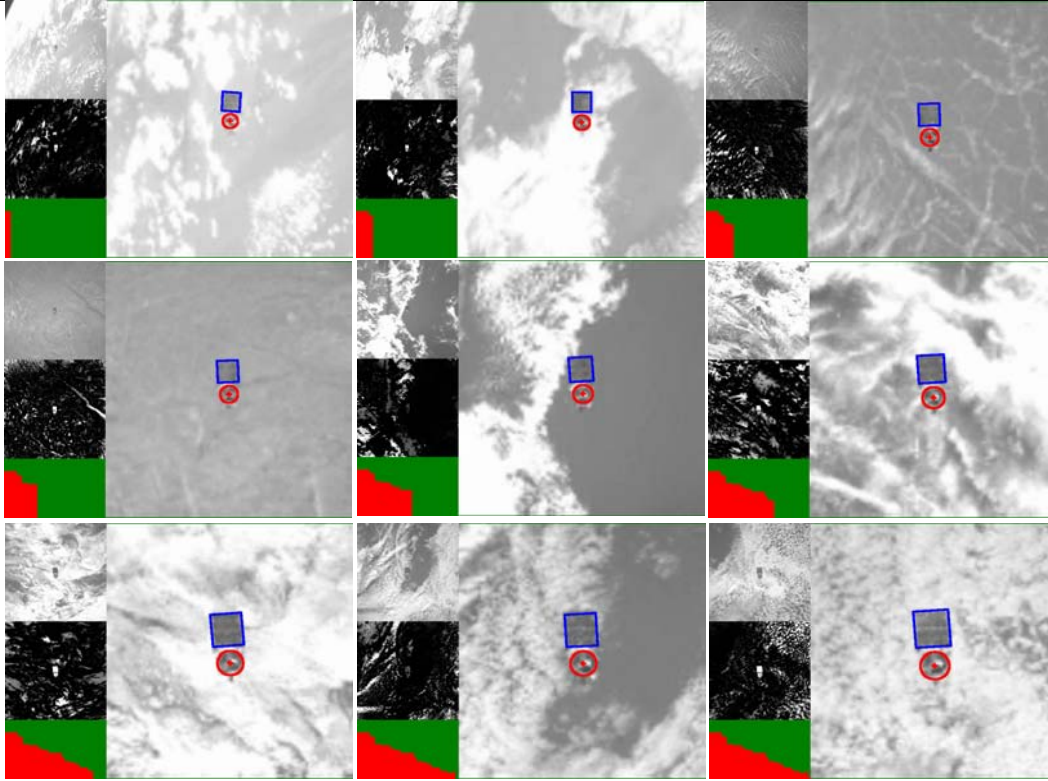

Fig. 11. Results of the model-based image tracking in an Orbital Express sequence.

**Simulated RV trajectories using image-based navigation**

SENER is currently implementing a generic simulator for the rendez-vous and docking manoeuvre to validate the integration of the data provided by the described image processing module with the control laws and procedures designed to guide the manoeuvre. In fig. 12 is presented a diagram of the simulator, including modules to describe the spacecraft dynamics, sensors, Kalman filtering stage [10, 11], actuators and AOCS/GNC control and guidance laws.

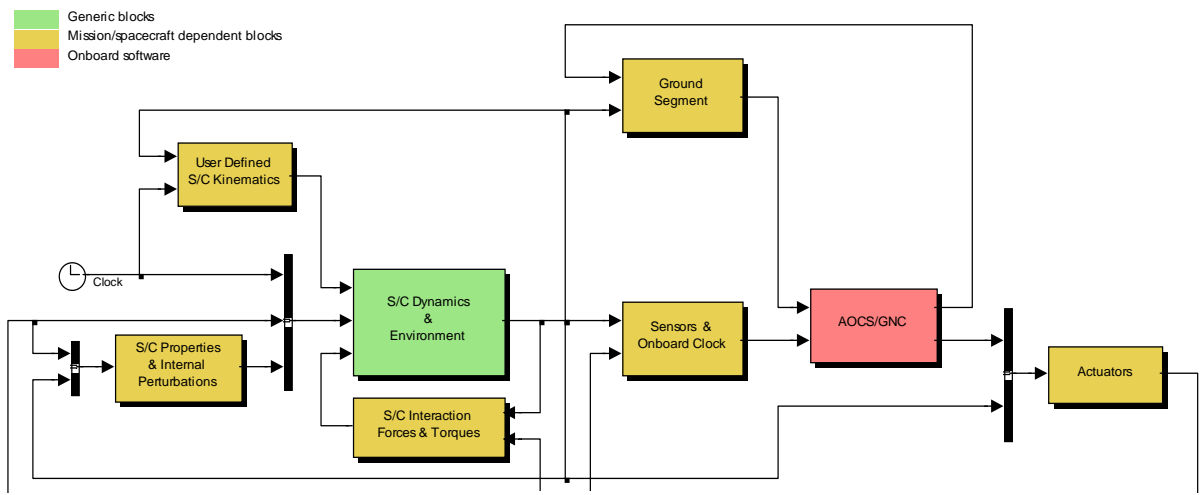

Fig. 12. Diagram of the rendez-vous and docking generic simulator

Two trajectories have been preliminarily simulated. The first one is a forced translation along V-bar from +2000m of the client to +100m. This trajectory is processed through two tangential boosts (one at the beginning and one at the end of the translation) and a continuous radial boost.

The second trajectory is a fly-around to pass from the V-bar axis to -50m on R-bar axis. This maneuver is performed with two tangential (V-bar) boosts at the beginning and at the end of the maneuver. The corresponding initial point is +118m on V-Bar.

Figure 13 (left panel) shows the measurement error (measure – real state) compared to the estimation error (estimated state – real state) and the covariance (error estimated by the filter) for the forced translation. This trajectory is quite short (about 15 min). This graph shows that the Kalman filter allows reducing the measurement error. Furthermore, the covariance stays higher than the error, ensuring that the filter provides a realistic precision of its estimation. The performances at the end of the trajectory could be improved dividing the trajectory into two parts. This will allow filter parameters to be adapted more precisely to the distance between both spacecraft and to the measurement profile, giving more robustness and precision to the estimation.

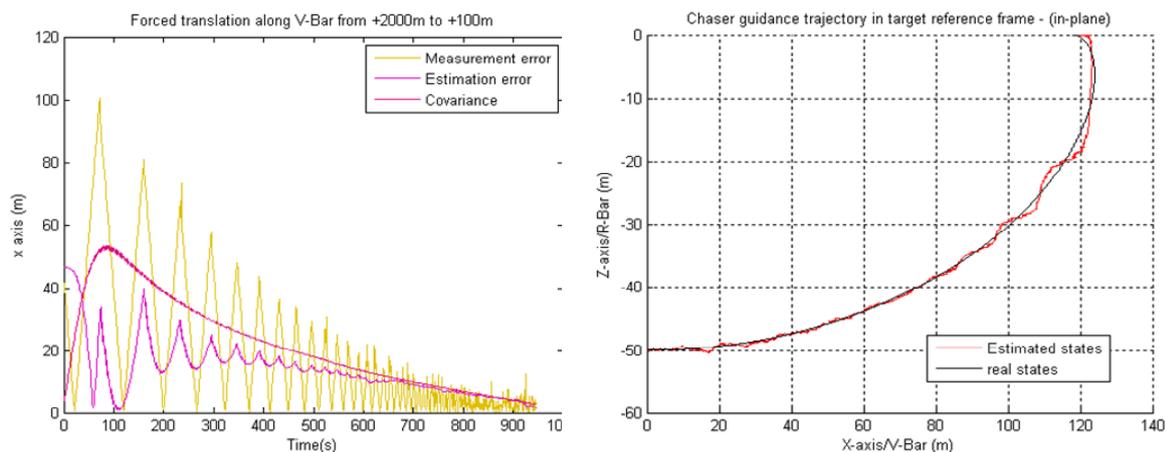

Fig. 13. Quantitative performances of the two simulated trajectories (see text).

The right panel of fig. 13 represents the fly-around trajectory in the reference frame tied to the target. The black line is the ideal states and the red one corresponds to the estimation. This trajectory presents the advantage that the distance between both spacecrafts is smaller and changes only slightly (from about 118m to 50m). The camera profile is more regular and the discontinuities are smaller so that the filter estimates easier the states. The error of the estimation at the end of the simulation was less than two meters which is a acceptable performance at this stage of the rendez-vous.

Extensive simulation of all mission phases will be performed during the following months to verify the accuracy and reliability of the developed system.

**CONCLUSIONS**

Automated safe rendez-vous is a primary objective for deep space exploration, technological update in-orbit or life extension of existing satellites. There have been some previous and successful tries in rendez-vous and docking, but none of them completely adaptable to its use on non cooperative targets. In this paper it has been outlined the image processing procedure developed by SENER to use a passive CCD camera as a rendez-vous relative navigation sensor, capable of operating with client satellites not equipped with any specific navigational aid to provide support during the approach. Extensive simulation of all rendez-vous phases with a definite target satellite will be performed during the following months to verify the accuracy and reliability of the developed system for the OLEV mission.

**REFERENCES**

[1]     P. Soille, *Morphological Image Analysis*, 2º Ed., Springer-Verlag, Berlin, 2004.

[2]     E. R. Dougherty and R. A. Lotufo, *Hands-on Morphological Image Processing*, SPIE Press, Washington, 2003.

[3]     P. Salembier and J. Serra, "Flat Zones filtering, Connected Operators, and Filters by Reconstruction", *IEEE Transactions on Image Processing*, vol. 4 (8), pp. 1153-1160, 1995.

[4]     L. Vincent, "Morphological Grayscale Reconstruction in Image Analysis: Applications and Efficient Algorithms", *IEEE Transactions on Image Processing*, vol. 2 (2), pp. 176-201, 1993.

[5]  A. Banerji and J. Goutsias, "A Morphological Approach to Automatic Mine Detection Problems", *IEEE Transactions on Aerospace and Electronic Systems*, vol. 34 (4), pp. 1085-1096, 1998.

[6]  M. Sezgin, and B. Sankur, "Survey over image thresholding techniques and quantitative performance evaluation, *Journal of Electronic Imaging*, vol. 13 (1), pp. 146–165, 2004.

[7]  R. M. Haralick and L.G. Shapiro, *Computer and Robot Vision*, vol. I, Addison-Welsey, Reading, Massachusetts, 1992.

[8]  Information on the ATV available on-line at: http://www.esa.int/esaMI/ATV/index.html.

[9]  J. Canny, "A Computational Approach To Edge Detection", *IEEE Transactions on Pattern Analysis and Machine Intelligence*, vol. 8, pp. 679-714, 1986.

[10] W. Fehse, *Automated Rendezvous and Docking of Spacecraft*, Cambridge Aerospace Series, Cambridge University Press, Cambridge, 2003.

[11] W. S. Levine (Ed.), *The Control Handbook*, The electrical engineering handbook series, CRS Press and IEEE Press, Florida, 1996.